\documentclass[11pt]{article}

\usepackage[final]{latex/acl}

\usepackage{times}
\usepackage{latexsym}
\usepackage[T1]{fontenc}
\usepackage[utf8]{inputenc}
\usepackage{microtype}
\usepackage{inconsolata}
\usepackage{graphicx}
\usepackage{booktabs}
\usepackage{amsmath}
\usepackage{xcolor}
\usepackage{colortbl}
\usepackage{multirow}
\usepackage{arydshln}

\newcommand{\method}{CoFiE}
\newcommand{\ub}[1]{\textcolor{gray}{#1}}

\title{\method{}: Coarse-to-Fine Evidence Selection\\ for Efficient Streaming Video Understanding}

\author{
Jing Jiang\textsuperscript{1,2}\thanks{Equal contribution.},
Yiran Ling\textsuperscript{1,2}\footnotemark[1],
Ruonan Li\textsuperscript{3},
Dimitrios Stamoulis\textsuperscript{1,2}\thanks{Corresponding authors.},
Jie Liu\textsuperscript{1,2}\footnotemark[2] \\
\textsuperscript{1}Harbin Institute of Technology \\
\textsuperscript{2}State Key Laboratory of Smart Farm Technologies and Systems \\
\textsuperscript{3}Pengcheng Laboratory \\
\texttt{\{25S003007, 25B903029\}@stu.hit.edu.cn,} \\
\texttt{lirn@pcl.ac.cn, \{dimi, jieliu\}@hit.edu.cn}
}

\begin{document}
\maketitle

\begin{abstract}

Streaming video understanding requires Vision Language Models (VLLMs) to process growing video streams and answer user questions under tight latency constraints. Existing methods improve efficiency with token pruning and memory-bank schemes, but mainly reduce visual tokens \textit{after} the LLM encoding stage. Therefore, downstream token pruning alone cannot substantially reduce end-to-end latency, since the expensive frame encoding cost has already been incurred. In this work, we propose CoFiE, a \textbf{Co}arse-to-\textbf{Fi}ne \textbf{E}vidence selection framework, that decouples evidence selection into a \textit{coarse} query-agnostic filtering stage before the vision encoder and a \textit{fine} query-specific refinement stage during LLM prefill. We introduce two lightweight modules, namely the coarse Novelty-Guided Frame Filtering and the fine Query-Specific Evidence Refinement, that retain visually distinctive candidates and refine the most relevant frames, respectively. This design removes substantial redundancy before frame encoding while retaining query-specific refinement once semantic information becomes available. Experimental results show that CoFiE establishes a new state-of-the-art accuracy--efficiency trade-off with open-source VLLMs across multiple video understanding benchmarks, reaching 78.86\% on StreamingBench and 68.72\% on OvO-Bench, improving over prior best methods by up to 3.15\%. Notably, even under aggressive evidence-frame filtering up to 80\%, CoFiE outperforms open-source frontier multimodal models while improving end-to-end inference latency by up to 2.54$\times$. 

\end{abstract}

\section{Introduction}

While frontier Vision Large Language Models (VLLMs) already achieve strong performance on offline video question answering~\citep{bai2025qwen3vl,wang2024qwen2vl,li2024llava}, streaming long-video understanding remains a distinct challenge~\citep{yao2025timechatonline}. Unlike offline settings, where the VLLM receives the fixed clip and user task before answering, streaming models must process a continuously growing visual stream and respond to timestamped user queries~\citep{li2025ovobench}. Consequently, query-specific evidence can only be identified \textit{after} the question has been processed together with the visual tokens~\citep{qian2025dispider}. This makes offline efficiency techniques less effective in streaming settings, where early reduction may discard short-lived events that become critical for later questions under strict latency constraints~\cite{zeng2025streamforest}.

\begin{figure}
    \centering
    \includegraphics[width=0.99\linewidth]{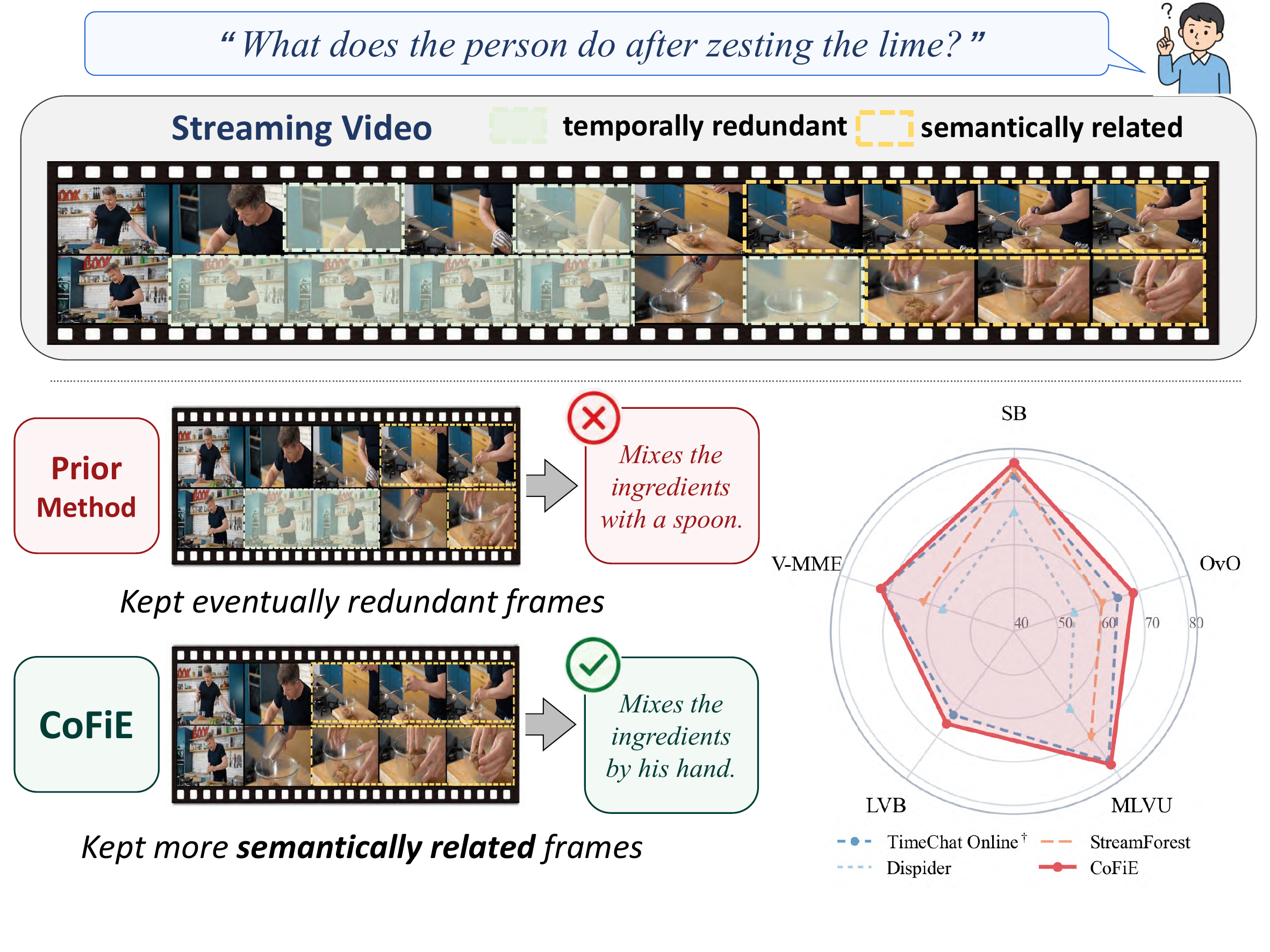}
    \caption{CoFiE presents a lightweight approach to retain semantically relevant and filter visually redundant frames, establishing state-of-the-art performance across multiple video understanding benchmarks, improving over prior best methods by up to 3.15\%.}
    \label{fig:teaser}
\end{figure}

Recent online acceleration methods introduce complex compression~\citep{yao2025timechatonline} and memory-bank schemes~\citep{zeng2025streamforest,zhang2024flashvstream} to reduce visually redundant content. However, existing methods perform evidence reduction \textit{after} frame encoding has already occurred~\citep{zhang2024flashvstream}, so the expensive encoding computation is paid in full regardless of how aggressively visual tokens are pruned downstream. Moreover, these approaches rely on nearly uniform fixed-rate sampling or compression rules, despite the fact that evidence in streaming videos is highly non-uniform and may include rare but answer-critical events~\citep{li2025ovobench}. As a result, while these methods often improve efficiency and accuracy, their end-to-end gains remain limited compared with unmodified baseline VLLMs that process denser visual context (Figure~\ref{fig:teaser}).

To overcome these limitations, we introduce CoFiE, a \textbf{Co}arse-to-\textbf{Fi}ne \textbf{E}vidence Selection framework, which decouples streaming evidence selection into two stages, a \textit{coarse} query-agnostic stage and a \textit{fine} query-specific stage. Our key intuition is to first keep a high-recall set of visually distinctive candidate frames, and then rank these candidates into evidence frames that are important for the user question. In the coarse stage before the visual encoder, a Novelty-Guided Frame Filtering (NGFF) module assigns each frame a histogram-based novelty score, retaining visually distinctive frames as high-recall candidates while discarding redundant ones. In the fine stage, a Query-Specific Evidence Refinement (QER) module ranks the retained candidates by their relevance to the question semantics using text-to-visual attention at LLM prefill. Our lightweight method allows only a compact set of high-recall candidate frames to interact with the text query, providing only question-relevant evidence frames for the LLM decoding stage, improving both end-to-end latency and downstream performance. 

Our comprehensive evaluation across state-of-the-art streaming video understanding benchmarks shows that CoFiE outperforms previously best methods by up to 3.15\% (Figure~\ref{fig:teaser}). We summarize our contributions as follows:
\begin{itemize}
    \item We propose CoFiE, a coarse-to-fine evidence selection framework for efficient streaming video understanding. To the best of our knowledge, CoFiE is the \textbf{first} framework to perform pre-encoder frame-level pruning for streaming VLLM inference.
    \item We decouple evidence selection into a query-agnostic coarse stage and a query-specific fine stage, implemented as two lightweight inference-time components. NGFF uses pre-encoder visual novelty to retain high-recall candidate frames, while QER aggregates text-to-visual attention at the frame level to select question-relevant evidence during prefill.
    \item We evaluate CoFiE with extensive experiments and ablation studies and we demonstrate state-of-the-art performance across multiple streaming and long-video understanding benchmarks.
\end{itemize}

\begin{figure*}[t]
\centering
\includegraphics[width=0.95\textwidth]{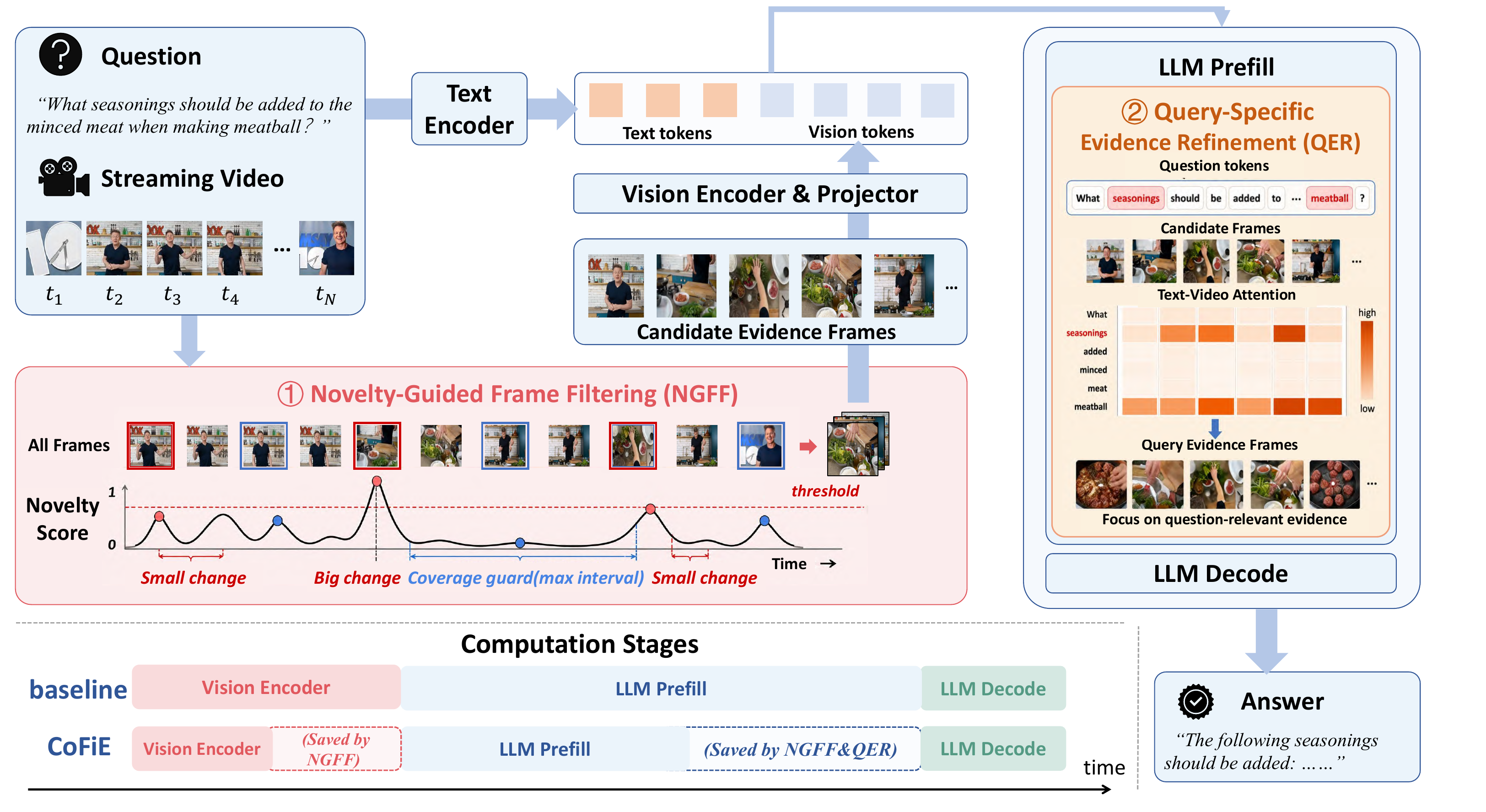}
\caption{Method overview. \method{} first applies Novelty-Guided Frame Filtering (NGFF) before visual encoding to keep a compact set of non-redundant candidate frames. It then applies Query-Specific Evidence Refinement (QER) during multimodal prefill to rank the encoded candidate frames with text-to-video attention and retain the query-specific evidence frames for answer generation.}
\label{fig:method}
\end{figure*}

\section{Related Work}

Streaming video understanding has become a central capability for frontier VLLMs~\cite{yao2025timechatonline}.
Open model families such as InternVL ~\citep{chen2023internvl}, LLaVA-OneVision ~\citep{li2024llava}, and Qwen-VL ~\citep{wang2024qwen2vl,bai2025qwen25vl,bai2025qwen3vl}, together with proprietary models such as GPT, Claude, and Gemini, report strong multimodal video performance and continue to narrow the gap to human-level results. However, these gains come with substantial inference cost: processing long inputs in streaming video understanding requires many visual tokens~\citep{chen2024videollmonline}, as context accumulates over time and user queries may arrive before the video ends~\citep{zhang2024flashvstream}. VLLMs must answer under partial observation, preserve useful history, and remain responsive as new frames arrive~\citep{yao2025timechatonline}. 

Several methods therefore aim to reduce the visual context in VLLMs, either by sampling fewer frames, compressing streaming memory, or pruning visual tokens. Memory-based methods such as VideoStreaming~\citep{qian2024videostreaming}, VideoLLM-online~\citep{chen2024videollmonline}, Flash-VStream~\citep{zhang2024flashvstream}, StreamForest~\citep{zeng2025streamforest}, and Dispider~\citep{qian2025dispider}, maintain compact video representations through various memory instantiations, including persistent event banks, memory-augmented retrieval, and disentangled perception and decision events. TimeChat-Online~\citep{yao2025timechatonline} removes redundant streaming tokens based on temporal visual change, while GlimpsePrune~\citep{zeng2025glimpseprune} and PruneVid~\citep{huang2024prunevid} use dynamic or question-aware pruning to reduce visual tokens during inference. 

Overall, existing approaches show that streaming videos contain substantial redundancy~\citep{yao2025timechatonline}, but they also expose a remaining challenge: fixed compression ratios and single-stage reduction rules can be brittle in online video understanding, where the relevant evidence depends on both temporal change and the user query. Unlike these methods, \method{} explicitly separates evidence selection from query-specific evidence refinement, allowing early visual reduction to remain cheap while deferring answer-specific selection until the user query is available.

\section{Methodology}

\subsection{Problem Formulation \& \method{} Overview}

In online video understanding, the VLLM first encodes the video frames into visual tokens, and then uses these tokens together with the user question during multimodal prefill and answer generation~\citep{bai2025qwen3vl}. Formally, let $V_{1:T}=\{I_1,\ldots,I_T\}$ be the frames sampled from the current video prefix, and let $q$ be the user question. The standard inference path is
\begin{equation}
X=\phi(V_{1:T}),
\qquad
y \sim p_\theta(y \mid X, q),
\end{equation}
where $\phi$ is the visual encoder, $X$ is the resulting visual-token sequence, and $y$ is the generated answer. Reducing frames before the encoder $\phi$ saves the cost of visual encoding and projection, while reducing tokens after $\phi$ mainly shortens the multimodal context used in prefill and decoding.
However, \textit{before} visual encoding, the model can only use cheap and query-agnostic signals from the raw frames, such as visual frame changes; query-specific evidence becomes available only after the question is processed with the visual tokens, when the retained frames have already been encoded.

\paragraph{Method Overview} \method{} views the inference path as a \textit{coarse-to-fine} selection problem comprising two stages, as illustrated in Figure~\ref{fig:method}. First, in the \textit{coarse} stage before visual encoding, we aim to keep a compact set of frames that may contain visual evidence, including gradual changes that accumulate over time, without deciding which frames answer the question. This is achieved based on a lightweight novelty-scoring mechanism in the Novelty-Guided Frame Filtering (NGFF) module, which returns a set of coarse selected candidate frames $S_c$. After these frames are encoded with the question in the text-to-visual attention computation during LLM prefill, \method{} introduces a ranking scheme in the Query-Specific Evidence Refinement (QER) module to refine these candidates into question-specific evidence frames, yielding an even smaller set of frames $S_e$ that reflect the frames most useful for downstream decoding and answer generation. We implement \method{} on top of Qwen3-VL-8B-Instruct~\citep{bai2025qwen3vl}. Next, in  Sections~\ref{sec:ngff} and~\ref{sec:qer}, we describe in detail the two main stages of our method.

\subsection{Novelty-Guided Frame Filtering}
\label{sec:ngff}

NGFF receives the sampled frames $V_{1:T}$ and returns the candidate-frame set $S_c$ before visual tokens are produced. At this point, the selector must be available before visual encoding, independent of the question, and cheaper than frame scoring with external vision-language models. This rules out heavier strategies based on auxiliary importance models, feature extraction, or motion estimation which would introduce overhead comparable to the visual encoder~\citep{qian2025dispider, yao2025timechatonline}. We therefore use a heuristic novelty score for candidate-frame selection based on grayscale \textit{histogram change}, as it requires no additional model or training, and is \textit{less} sensitive to small spatial shifts than pixel-wise difference~\citep{histogram2013}.

\paragraph{Lightweight histogram-based novelty.}
For each sampled frame $I_t$, NGFF computes a compact grayscale histogram: first, we convert RGB values to grayscale intensity, and we
then form a normalized $B$-bin histogram vector $h_t$ over the normalized intensity range of pixel values, i.e., $\sum_b h_t[b]=1$.
We measure the difference between two frames $i$ and $j$ as the total-variation distance:
\begin{equation}
d(h_i,h_j)=\frac{1}{2}\|h_i-h_j\|_1
\end{equation}

To capture visually novel frames, for each intermediate sampled frame $I_t$ with histogram $h_t$, NGFF compares its distance with \textbf{both} the immediately previous sampled frame and the most recent candidate frame at index $r(t)$:
\begin{equation}
s_t^{\mathrm{adj}} = d(h_t,h_{t-1}),
\qquad
s_t^{\mathrm{ref}} = d(h_t,h_{r(t)}),
\end{equation}
\begin{equation}
s_t^{\mathrm{NGFF}}=\max(s_t^{\mathrm{adj}},s_t^{\mathrm{ref}}).
\end{equation}
Intuitively, the adjacent term captures abrupt local changes, while the candidate-reference term captures accumulated drift from the last candidate visual state.
Taking the maximum allows either type of change to trigger candidate-frame retention.

\paragraph{Gap-aware frame retention} To keep the candidate sequence well covered, we introduce temporal \textit{safeguards}: first, we always retain the first and last sampled frames. Moreover, if novelty-based selection returns fewer than $M_{\min}$ candidates, we add uniformly spaced sampled frames in the candidate set. Last, for each $1<t<T$, NGFF retains frame $I_t$ as a candidate frame when either its novelty exceeds a threshold or the distance from the previous candidate frame becomes too large:
\begin{equation}
I_t \in S_c
\quad \text{IF} \quad
s_t^{\mathrm{ngff}}\ge \tau_h
\quad \text{OR} \quad
t-r(t)\ge L_{\max}.
\end{equation}
with $L_{\max}$ measured in sampled-frame indices rather than wall-clock time. Together, novelty-based retention and temporal safeguards produce the candidate-frame set $S_c=\{f_1,\ldots,f_m\}$; only these $m$ frames are sent to the encoder.


\subsection{Query-Specific Evidence Refinement}
\label{sec:qer}


At this stage, only the candidate frames selected by NGFF are encoded:
$X_c=\phi(\{I_t:t\in S_c\})$. The key intuition is that, once these frames are processed \textbf{with} the question, they can be scored by how strongly they support answer generation. QER performs this refinement during multimodal prefill, using text-to-visual attention to select the evidence frames from the candidate set. To implement this refinement, we use the frame-token structure provided by Qwen3-VL-style inputs~\citep{bai2025qwen3vl}. Each candidate frame is enclosed by vision boundary tokens and mapped to a group of visual tokens with temporal, height, and width positions~\citep{bai2025qwen25vl}.

Let $\mathcal{V}_f$ denote the visual-token indices corresponding to frame $f$, and let $\mathcal{Q}$ denote the textual-token indices used for scoring, i.e., all non-visual tokens in the prefill sequence.
At this stage, in middle-to-late decoder layers, attention scores  between query tokens and vision tokens provide a stable text-to-vision relevance signal~\citep{huang2024prunevid, chen2024fastv}. To this end, the QER module leverages the attention scores available at the last-layer attention map $\ell$ to compute compatibility between the user query and video frames, without introducing any additional computational overhead. Formally, for textual token $i$ and visual token $j$, we write:
\begin{equation}
a^{\ell}_{ij}
=
\frac{1}{H}
\sum_{h=1}^{H}
\frac{
\langle q^{\ell,h}_{i}, k^{\ell,h}_{j}\rangle
}{\sqrt{d_h}},
\end{equation}
where $H$ is the number of attention heads and $d_h$ is the head dimension. To achieve a lightweight implementation and avoid materializing a full text-by-vision attention matrix, QER computes these scores in tiles of size 512~\citep{chen2024fastv}.

\paragraph{Max-over-text aggregation} Next, we aggregate token-level scores into a frame-level relevance score. For each visual token, we first take the strongest textual association, then average over the frame's visual tokens:
\begin{equation}
w_f
=
\frac{1}{|\mathcal{V}_f|}
\sum_{j\in\mathcal{V}_f}
\max_{i\in\mathcal{Q}} a^{\ell}_{ij}.
\label{re:max}
\end{equation}
We emphasize here the use of the maximum operator: averaging attention over all textual tokens may dilute localized associations, especially when the question-relevant object occupies only a small portion of the frame. Therefore, our max-over-text aggregation remains recall-oriented at the frame level: a frame should be retained if any of its visual regions is strongly associated with a question token, ensuring that Equation~\ref{re:max} robustly identifies relevant frames even if they relate to a small visual region or a single queried object.

\paragraph{Frame refinement}
Given a keep ratio $\rho$, QER retains the top-scoring candidate frames:
\begin{equation}
\begin{aligned}
k &= \max(1,\operatorname{round}(\rho |S_c|)),\\
S_e &= \operatorname{TopK}_{f\in S_c}(w_f,k).
\end{aligned}
\end{equation}
Frames outside $S_e$ are removed from the hidden states, positional embeddings, and compatible cached states before the remaining layers continue. The retained evidence frames are restored to chronological order, and the model completes prefill and generation on the shortened sequence. Scores are computed in chunks during the forward pass to avoid materializing the full text-by-vision attention matrix. We note that $\rho$ is \textbf{not} a bespoke method parameter: following standard inference evaluations, it defines the target computation budget (relative to a full-model execution), allowing representative comparisons with state-of-the-art approaches~\citep{yao2025timechatonline}.

\begin{table*}[t]
\centering
\scriptsize
\setlength{\tabcolsep}{3.2pt}
\renewcommand{\arraystretch}{1.06}
\resizebox{0.9\textwidth}{!}{
\begin{tabular}{l!{\vrule width 0.3pt}cccccccccc!{\vrule width 0.3pt}c}
\toprule
\multirow{2}{*}{\textbf{Method}} & \multicolumn{11}{c}{\textbf{StreamingBench}} \\
\cmidrule(lr){2-12}
 & \textbf{OP} & \textbf{CR} & \textbf{CS} & \textbf{ATP} & \textbf{EU} & \textbf{TR} & \textbf{PR} & \textbf{SU} & \textbf{ACP} & \textbf{CT} & \textbf{Acc.} \\
\midrule
\rowcolor{black!10}
\multicolumn{12}{c}{\textit{Proprietary VLLMs}} \\
\ub{Gemini 1.5 Pro}~\citep{google2024gemini15} & \ub{79.0} & \ub{80.5} & \ub{83.5} & \ub{79.7} & \ub{80.0} & \ub{84.7} & \ub{77.8} & \ub{64.2} & \ub{72.0} & \ub{48.7} & \ub{75.7} \\
\ub{GPT-4o}~\citep{openai2024gpt4o} & \ub{77.1} & \ub{80.5} & \ub{83.9} & \ub{76.5} & \ub{70.2} & \ub{83.8} & \ub{66.7} & \ub{62.2} & \ub{69.1} & \ub{49.2} & \ub{73.3} \\
\ub{Claude 3.5 Sonnet}~\citep{anthropic2024claude35sonnet} & \ub{80.49} & \ub{77.34} & \ub{82.02} & \ub{81.73} & \ub{72.33} & \ub{75.39} & \ub{61.11} & \ub{61.79} & \ub{69.32} & \ub{43.09} & \ub{72.44} \\
\midrule
\rowcolor{black!10}
\multicolumn{12}{c}{\textit{Open-Source Offline Video VLLMs}} \\
MiniCPM-V~\citep{yao2024minicpmv} 2.6 & 71.93 & 71.09 & 77.92 & 75.82 & 64.60 & 65.73 & 70.37 & 56.10 & 62.32 & 53.37 & 67.44 \\
InternVL-V2~\citep{chen2023internvl} & 68.12 & 60.94 & 69.40 & 77.12 & 67.70 & 62.93 & 59.26 & 53.25 & 54.96 & 56.48 & 63.72 \\
VILA-1.5~\citep{lin2024vila} & 53.68 & 49.22 & 70.98 & 56.86 & 53.42 & 53.89 & 54.63 & 48.78 & 50.14 & 17.62 & 52.32 \\
Video-LLaMA2~\citep{cheng2024videollama2} & 55.86 & 55.47 & 57.41 & 58.17 & 52.80 & 43.61 & 39.81 & 42.68 & 45.61 & 35.23 & 49.52 \\
LLaVA-OneVision~\citep{li2024llava} & 80.38 & 74.22 & 76.03 & 80.72 & 72.67 & 71.65 & 67.59 & 65.45 & 65.72 & 45.08 & 71.12 \\
Qwen2-VL-7B~\citep{wang2024qwen2vl} & 75.2 & 82.81 & 73.19 & 77.45 & 68.32 & 71.03 & 72.22 & 61.19 & 61.47 & 46.11 & 69.04 \\
Qwen2.5-VL-7B~\citep{bai2025qwen25vl} & 78.32 & 80.47 & 78.86 & 80.45 & 76.73 & 78.50 & 79.63 & 63.41 & 66.19 & 53.19 & 73.68 \\
\hdashline
Qwen3-VL-8B~\citep{bai2025qwen3vl} & 81.84 & 80.47 & 82.97 & 83.01 & 75.47 & 82.87 & 81.48 & 65.04 & 67.05 & 53.19 & 75.88 \\
Qwen3-VL-8B (drop 50\%) & 81.57 & 81.25 & 81.70 & 83.97 & 76.73 & 82.24 & 80.56 & 65.85 & 69.03 & 54.79 & 76.28 \\
Qwen3-VL-8B (drop 80\%) & 79.40 & 79.69 & 79.50 & 82.37 & 71.70 & 76.64 & 81.48 & 65.04 & 63.64 & 54.79 & 73.56 \\
\midrule
\rowcolor{black!10}
\multicolumn{12}{c}{\textit{Open-source online video VLLMs}} \\
Dispider~\citep{qian2025dispider} & 74.92 & 75.53 & 74.10 & 73.08 & 74.44 & 59.92 & 76.14 & 62.91 & 62.16 & 45.80 & 67.63 \\ 
Flash-VStream~\citep{zhang2024flashvstream} & 25.89 & 43.57 & 24.91 & 23.87 & 27.33 & 13.08 & 18.52 & 25.20 & 23.87 & 48.70 & 23.23 \\ 
ViSpeak~\citep{fu2025vispeak} & 79.80 & 88.30 & 83.30 & 81.10 & 76.40 & 75.10 & 70.40 & 65.90 & 77.30 & 34.20 & 74.40 \\ 
StreamForest~\citep{zeng2025streamforest} & 83.11 & 82.81 & 82.65 & 84.26 & 77.50 & 78.19 & 76.85 & 69.11 & 75.64 & 54.40 & 77.26 \\ %
TimeChat-Online~\citep{yao2025timechatonline} & 80.22 & 82.03 & 79.50 & 83.33 & 76.10 & 78.50 & 78.70 & 64.63 & 69.60 & 57.98 & 75.36 \\ 
\hdashline
TimeChat-Online$^\dagger$ & 81.84 & 80.47 & 82.97 & 83.01 & 75.47 & 82.87 & 81.48 & 65.04 & 67.05 & 53.19 & 75.88 \\ 
TimeChat-Online$^\dagger$ (drop 50\%) & 80.23 & 82.11 & 78.80 & 81.87 & 78.30 & 75.54 & 75.47 & 63.39 & 63.96 & 54.44 & 72.53 \\
TimeChat-Online$^\dagger$ (drop 80\%) & 79.46 & 78.05 & 79.60 & 81.35 & 76.42 & 70.82 & 77.36 & 61.61 & 61.26 & 55.56 & 71.14 \\
\midrule
\textbf{CoFiE} & 81.03 & 84.38 & 90.54 & 83.92 & 74.68 & 81.31 & 88.89 & 67.07 & 73.30 & 58.51 & \underline{78.58} \\
\rowcolor{blue!9}
\textbf{CoFiE} (drop 50\%) & 82.93 & 84.38 & 89.91 & 82.96 & 75.32 & 84.11 & 87.96 & 67.48 & 72.44 & 57.45 & \textbf{78.86} \\
\textbf{CoFiE} (drop 80\%) & 82.93 & 83.59 & 88.96 & 83.28 & 74.68 & 79.44 & 87.96 & 65.85 & 72.44 & 55.85 & 77.82 \\
\bottomrule
\end{tabular}
}
\caption{Main results on StreamingBench (OP: Object Perception, CR: Causal Reasoning, CS: Clips Summarization, ATP: Attribute Perception, EU: Event Understanding, TR: Text-Rich Understanding, PR: Prospective Reasoning, SU: Spatial Understanding, ACP: Action Perception, CT: Counting, Acc.: average performance). The best two results, excluding the proprietary upper-bound models in \ub{gray}, are \textbf{bold} and \underline{underlined}, respectively. The drop 50\% setting means that 50\% of the original frames are removed; unless otherwise marked, all other rows are evaluated using 100\% of the video frames. TimeChat-Online$^\dagger$ denotes our improved reimplementation of TimeChat-Online with frontier VLLM updates for a representative comparison with CoFiE.}
\label{tab:streamingbench-main-results}
\end{table*}

\section{Experiments}
\label{sec:results}

\subsection{Experimental Setup}
\label{sec:experimental-setup}

\paragraph{Benchmarks.}
We evaluate \method{} across \textbf{both} \textit{streaming} video understanding and \textit{offline} long-video understanding benchmarks.
The streaming setting includes StreamingBench~\citep{lin2024streamingbench} and OvO-Bench~\citep{li2025ovobench}, while the offline setting includes MLVU~\citep{zhou2025mlvu}, LongVideoBench~\citep{wu2024longvideobench}, MVBench~\citep{li2024mvbench}, and Video-MME~\citep{fu2025videomme}. We report the standard evaluation metrics from each benchmark suite, along with inference time, to quantify the accuracy--efficiency trade-off.

\paragraph{Drop-ratio definition.}
The reported drop ratio is the proportion of the original input frames removed after both NGFF and QER. If NGFF removes a fraction $D_{\mathrm{NGFF}}$ of the original frames and QER subsequently removes a fraction $D_{\mathrm{QER}}$ of the NGFF-retained candidates, the total drop ratio is defined as:
\begin{equation}
D_{\mathrm{total}}=1-(1-D_{\mathrm{NGFF}})(1-D_{\mathrm{QER}}).
\end{equation}

\begin{table*}[t]
\centering
\tiny
\setlength{\tabcolsep}{3.6pt}
\renewcommand{\arraystretch}{1.06}
\resizebox{0.78\textwidth}{!}{
\begin{tabular}{l!{\vrule width 0.3pt}cccccc!{\vrule width 0.3pt}c}
\toprule
\multirow{2}{*}{\textbf{Method}} & \multicolumn{7}{c}{\textbf{OvO-Bench Real-Time Visual Perception}} \\
\cmidrule(lr){2-8}
 & \textbf{OCR} & \textbf{ACR} & \textbf{ATR} & \textbf{STU} & \textbf{FPD} & \textbf{OJR} & \textbf{Avg.} \\
\midrule
\midrule
\rowcolor{black!10}
\multicolumn{8}{c}{\textit{Proprietary VLLMs}} \\
\ub{Gemini 1.5 Pro}~\citep{google2024gemini15} & \ub{87.30} & \ub{67.00} & \ub{80.20} & \ub{54.50} & \ub{68.30} & \ub{67.40} & \ub{70.80} \\
\ub{GPT-4o}~\citep{openai2024gpt4o} & \ub{69.10} & \ub{65.10} & \ub{65.50} & \ub{50.00} & \ub{68.30} & \ub{63.70} & \ub{63.60} \\
\midrule
\rowcolor{black!10}
\multicolumn{8}{c}{\textit{Open-source offline video VLLMs}} \\
Qwen2-VL-7B~\citep{wang2024qwen2vl} & 69.13 & 53.21 & 63.79 & 50.56 & 66.34 & 60.87 & 60.65 \\
LLaVA-NeXT-Video-7B~\citep{li2024llava} & 69.80 & 59.60 & 66.40 & 50.60 & 72.30 & 61.40 & 63.30 \\
LLaVA-OneVision-7B~\citep{li2024llava} & 67.10 & 58.70 & 69.80 & 49.40 & 71.30 & 60.30 & 62.80 \\
InternVL-V2-8B~\citep{opengvlab2024internvl2} & 68.50 & 58.70 & 69.00 & 44.90 & 67.30 & 56.00 & 60.70 \\
LongVU-7B~\citep{shen2024longvu} & 55.70 & 49.50 & 59.50 & 48.30 & 68.30 & 63.00 & 57.40 \\
\hdashline
Qwen3-VL-8B~\citep{bai2025qwen3vl} & 77.85 & 59.63 & 73.28 & 53.37 & 69.00 & 57.07 & 65.03 \\
Qwen3-VL-8B (drop 50\%) & 79.19 & 61.47 & 71.55 & 55.62 & 68.00 & 57.61 & 65.57 \\
Qwen3-VL-8B (drop 80\%) & 75.17 & 56.88 & 71.55 & 50.00 & 67.00 & 59.78 & 63.40 \\
\midrule
\rowcolor{black!10}
\multicolumn{8}{c}{\textit{Open-source online video VLLMs}} \\
Dispider~\citep{qian2025dispider}  & 57.72 & 49.54 & 62.07 & 44.94 & 61.39 & 51.63 & 54.55 \\
TimeChat-Online~\citep{yao2025timechatonline} & 75.20 & 46.80 & 70.70 & 47.80 & 69.30 & 61.40 & 61.90 \\
Flash-VStream~\citep{zhang2024flashvstream} & 25.50 & 32.10 & 29.30 & 33.70 & 29.70 & 28.80 & 29.90 \\
StreamForest~\citep{zeng2025streamforest} & 68.46 & 53.21 & 71.55 & 47.75 & 65.35 & 60.87 & 61.20 \\
\hdashline
TimeChat-Online$^\dagger$ & 77.85 & 59.63 & 73.28 & 53.37 & 69.00 & 57.07 & 65.03 \\
TimeChat-Online$^\dagger$ (drop 50\%) & 78.52 & 58.72 & 73.28 & 51.12 & 66.00 & 59.24 & 64.48 \\
TimeChat-Online$^\dagger$ (drop 80\%) & 72.48 & 55.05 & 68.97 & 48.88 & 66.00 & 53.26 & 60.77 \\
\midrule
\textbf{CoFiE} & 79.87 & 58.72 & 78.45 & 56.18 & 75.00 & 60.33 & \underline{68.09} \\
\rowcolor{blue!9}
\textbf{CoFiE} (drop 50\%) & 80.54 & 59.63 & 76.72 & 56.74 & 74.00 & 64.67 & \textbf{68.72} \\
\textbf{CoFiE} (drop 80\%) & 76.51 & 52.29 & 75.86 & 53.37 & 76.00 & 61.96 & 66.00 \\
\bottomrule
\end{tabular}
}
\caption{Main results on the OvO-Bench Real-Time Visual Perception category (OCR: Optical Character Recognition, ACR: Action Recognition, ATR: Attribute Recognition, STU: Spatial Understanding, FPD: Future Prediction, OJR: Object Recognition, Avg.: average performance). The best two results, excluding the proprietary upper-bound models in \ub{gray}, are \textbf{bold} and \underline{underlined}, respectively. The drop 50\% setting means that 50\% of the original frames are removed; unless otherwise marked, all other rows are evaluated using 100\% of the video frames.}
\label{tab:ovo-bench-main-results}
\end{table*}

\subsection{Main Results: Online Benchmarks}

Table~\ref{tab:streamingbench-main-results} and Table~\ref{tab:ovo-bench-main-results} present the main comparisons on StreamingBench and the real-time visual perception category of OvO-Bench, respectively. Proprietary VLLMs are included as upper-bound references, while our comparisons focus on open-source offline and online video VLLMs. On StreamingBench, \method{} achieves 78.58\% accuracy without frame dropping and 78.86\% accuracy with 50\% frame dropping, outperforming the strongest prior open-source baseline StreamForest by 1.32\% and 1.60\%, respectively.

Compared with the Qwen3-VL-8B-Instruct backbone, \method{} improves the overall accuracy by up to 2.98\%. We emphasize that, to ensure a representative comparison, we also re-implement the strongest prior online baseline, namely TimeChat-Online~\citep{yao2025timechatonline}, under the same updated VLLM backbone used by CoFiE, and we fully reproduce the results. We denote this strengthened baseline as TimeChat-Online$^\dagger$ in Table~\ref{tab:streamingbench-main-results} and Table~\ref{tab:ovo-bench-main-results}. This setting ensures that we assess CoFiE against prior streaming-token reduction methods without an advantage from frontier VLLM capability, with CoFiE still outperforming the improved TimeChat-Online$^\dagger$ by up to 2.7\%.


Similarly, on OvO-Bench (Table~\ref{tab:ovo-bench-main-results}), \method{} reaches 68.72\% average score under the 50\% drop setting, surpassing the Qwen3-VL-8B-Instruct backbone by 3.69\% and outperforming all open-source offline and online baselines. Even when removing 80\% of the input frames, \method{} retains 99.0\% of its no-drop performance on StreamingBench and 96.9\% on OvO-Bench, suggesting that our coarse-to-fine selection robustly removes redundant frames while preserving important visual evidence.


\begin{figure}[h!]
\centering
\includegraphics[width=0.90\linewidth]{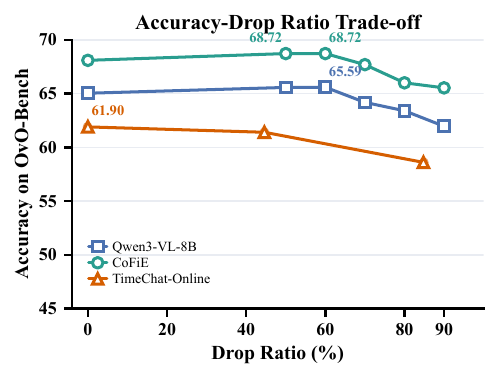}
\caption{Accuracy--drop trade-off on OvO-Bench. CoFiE achieves higher accuracy across drop ratios, even under aggressive frame dropping up to 90\%. The Qwen3-VL-8B curve applies NGFF and QER to the vanilla weights, while the CoFiE curve applies the same filters to the fine-tuned CoFiE checkpoint. }
\label{fig:accuracy-drop}
\end{figure}

\paragraph{Accuracy--Efficiency Trade-off.}

\begin{figure}[h]
\centering
\includegraphics[width=0.98\linewidth]{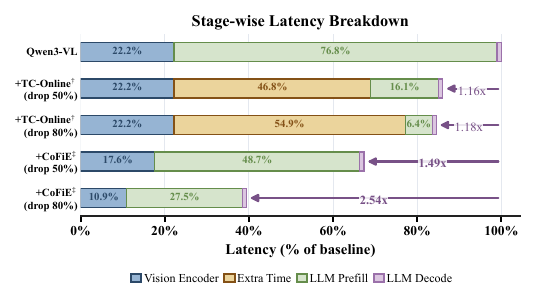}
\caption{Stage-wise latency breakdown on OvO-Bench, normalized by the Qwen3-VL-8B no-drop baseline. Stacked bars show vision encoding, extra selection time, LLM prefill, and LLM decoding.} 
\label{fig:stage-runtime}
\end{figure}

Figure~\ref{fig:accuracy-drop} illustrates the accuracy--drop ratio trade-off on OvO-Bench, comparing \method{} with the unpruned Qwen3-VL-8B backbone and TimeChat-Online. TimeChat-Online reaches its best accuracy of 61.90\% along the trade-off curve, whereas \method{} consistently dominates it across frame drop ratios. Notably, \method{} surpasses the full-frame baseline while processing substantially fewer frames, achieving its best accuracy of 68.72\% at a 50\% drop ratio and remaining robust at an aggressive 80\% drop ratio. Overall, \method{} improves Pareto performance by removing temporally redundant frames while preserving  query-specific visual evidence for online video understanding.


\begin{table}[t]
\centering
\tiny
\setlength{\tabcolsep}{1.8pt}
\renewcommand{\arraystretch}{1.05}
\resizebox{\columnwidth}{!}{
\begin{tabular}{l!{\vrule width 0.3pt}ccccc}
\toprule
\multirow{2}{*}{\textbf{Method}} & \multirow{2}{*}{\textbf{MLVU}} & \multirow{2}{*}{\textbf{LVB}} & \multirow{2}{*}{\textbf{MVB}} & \multicolumn{2}{c}{\textbf{Video-MME}} \\
\cmidrule(lr){5-6}
 & & & & \textbf{Long} & \textbf{All} \\ 
\midrule
Qwen2-VL-7B & -- & -- & 67.0 & -- & 63.3 \\ 
Qwen2.5-VL-7B & -- & -- & -- & 50.4 & 63.2 \\ 
Qwen3-VL-8B & 76.94 & 63.73 & 66.88 & 65.1 & 71.7 \\ 
\midrule
Dispider & 61.7 & -- & -- & -- & 57.2 \\ 
TimeChat-Online & 62.6 & 55.4 & -- & 48.4 & 62.4 \\ 
Flash-VStream & 66.3 & 42.0 & 65.4 & -- & -- \\ 
StreamForest & 70.0 & -- & 70.2 & -- & 61.9 \\ 
\midrule
\rowcolor{blue!9}
\method{} & \textbf{77.79} & \textbf{66.26} & \textbf{71.12} & \textbf{66.4} & \textbf{72.1} \\ 
\bottomrule
\end{tabular}
}
\caption{Offline benchmark results under the 0\% drop setting. LVB and MVB denote LongVideoBench and MVBench. Dashes indicate benchmarks for which the corresponding source papers did not report results.}
\label{tab:offline-results}
\end{table}

Figure~\ref{fig:stage-runtime} reports the stage-wise latency breakdown on OvO-Bench, normalized by the Qwen3-VL-8B no-drop baseline. For a fair comparison, all methods use the same backbone. \method{} adopts its training-free variant, while TimeChat-Online is reproduced by applying its core compression module to Qwen3-VL without additional fine-tuning. Compared with TimeChat-Online, which yields limited end-to-end speedups of 1.16$\times$ and 1.18$\times$ at the 50\% and 80\% drop settings, \method{} achieves 1.49$\times$ and 2.54$\times$, respectively. The gain mainly comes from reducing computation before visual encoding. Overall, \method{} delivers a substantially stronger end-to-end efficiency gain by targeting the dominant front-end visual computation.

\subsection{Offline Long-video Results}

For a comprehensive evaluation, we expand our methodology assessment to additional benchmarks: Table~\ref{tab:offline-results} reports the offline long-video results on MLVU, LongVideoBench, MVBench, and Video-MME under the 0\% drop setting.
Although \method{} is designed for efficient streaming video understanding, it also improves the unpruned Qwen3-VL-8B backbone across all reported offline benchmarks.
Against prior online video VLLMs, the gains are larger, especially on MLVU and LongVideoBench, where \method{} outperforms the strongest reported online baselines by 7.79\% and 10.86\%, respectively. These results demonstrate that our fine-tuned model improves long-video reasoning ability beyond the streaming setting, rather than merely optimizing for online latency.

\begin{table}[t]
\centering
\small
\setlength{\tabcolsep}{5.0pt}
\renewcommand{\arraystretch}{1.08}
\begin{tabular}{lccc}
\toprule
\textbf{Variant} & \textbf{Drop (\%)} & \textbf{Avg.} & \textbf{$\Delta$Avg.} \\
\midrule
Qwen3-VL-8B & 0.0 & 65.03 & -- \\
\midrule
NGFF-only & 20.0 & 65.57 & +0.54 \\
QER-only & 37.5 & 65.03 & +0.00 \\
\rowcolor{blue!8}
NGFF20\% + QER37.5\% & 50.0 & \textbf{66.85} & \textbf{+1.82} \\
\midrule
NGFF-only & 50.0 & 63.40 & -1.63 \\
QER-only & 60.0 & 65.03 & +0.00 \\
NGFF50\% + QER60\% & 80.0 & 64.95 & -0.08 \\
\midrule
NGFF-only & 60.0 & 61.96 & -3.07 \\
QER-only & 75.0 & 65.03 & +0.00 \\
NGFF60\% + QER75\% & 90.0 & 62.99 & -2.04 \\
\bottomrule
\end{tabular}
\caption{Components ablation of NGFF and QER on the Real-Time Visual Perception category of OvO-Bench. QER consistently improves NGFF-pruned variants by refining query-relevant visual evidence after the initial novelty-based filtering stage.}
\label{tab:ablation}
\end{table}

\begin{table}[t]
\centering
\small
\setlength{\tabcolsep}{4.4pt}
\renewcommand{\arraystretch}{1.08}
\resizebox{\columnwidth}{!}{
\begin{tabular}{lcc}
\toprule
\multirow{2}{*}{\textbf{Filtering signal}} & \textbf{Inference} & \multirow{2}{*}{\textbf{Accuracy}} \\
 & \textbf{time (s)} & \\
\midrule
\multicolumn{3}{l}{\textit{Baseline (drop 0\%)}} \\
Uniform Sampling & 319 & 65.03 \\
\midrule
\multicolumn{3}{l}{\textit{Drop 20\%}} \\
Low-Res Frame Difference~\citep{lipton1998moving} & 292 & 61.01 \\
dHash~\citep{zauner2010implementation} & 298 & 61.73 \\
Downsampled SSIM~\citep{wang2004image} & 297 & 62.54 \\
Optical Flow~\citep{horn1981determining} & 321 & 62.39 \\
Scene Change Detection + H-Diff~\citep{zhang1993automatic} & 294 & 61.01 \\
\midrule
\rowcolor{blue!8}
Histogram Difference & 293 & 63.58 \\
\bottomrule
\end{tabular}
}
\caption{Capturing the effect of different lightweight filtering signals in NGFF. Histogram Difference is the default signal used in \method{}.}
\label{tab:hdiff-signal}
\end{table}

\subsection{Ablation Studies}

\paragraph{NGFF and QER component ablation.}
Table~\ref{tab:ablation} isolates the effect of each stage by comparing NGFF-only and QER-only variants on the Real-Time Visual Perception category of OvO-Bench. NGFF-only improves the unpruned baseline by 0.54\% at 20\% drop, but its accuracy decreases with the drop ratio. This behavior suggests that novelty-based filtering efficiently removes redundant frames, but a query-agnostic signal becomes insufficient once the compression budget discards sparse answer evidence.

QER consistently shifts this trade-off by refining NGFF candidates with query-dependent attention. Adding QER to NGFF20\% raises the average score while increasing the total drop ratio to 50\%. Under heavier compression, QER improves NGFF50\% at 80\% total drop and NGFF60\% at 90\% total drop. The 80\% two-stage variant still retains 99.9\% of the no-drop backbone accuracy. These results show that NGFF and QER are complementary, with NGFF reducing front-end visual computation and QER preserving task-critical visual evidence before generation.

\paragraph{Filtering-signal ablation.}

Table~\ref{tab:hdiff-signal} compares lightweight pre-encoder filtering signals under the same 20\% drop ratio. Histogram Difference achieves the highest accuracy among all adaptive signals, reaching 63.58\%. It also yields an 8.2\% end-to-end speedup while retaining 96.9\% of the baseline accuracy. This controlled comparison further shows that the default signal should be simple, but not limited to raw pixel-level differences. Low-Res Frame Difference and dHash are efficient, but they underperform Histogram Difference respectively, suggesting that local differences and compact hashes are brittle proxies for task-critical visual evidence. Downsampled SSIM is more competitive, but still trails Histogram Difference. Optical Flow captures motion structure, but its motion-estimation overhead makes it slower than the no-drop baseline. Overall, Histogram Difference provides the best speed--accuracy trade-off among practical front-end signals, making it an appropriate NGFF criterion for filtering redundant frames before visual encoding.

\section{Limitations and Future Work}
CoFiE focuses on frame-level evidence selection and does not explicitly optimize KV-cache management. By removing redundant frames before visual encoding and during multimodal prefill, our approach reduces the number of visual tokens entering the LLM and can partially alleviate KV-cache growth in long video streams. However, the cache is not jointly scheduled, compressed, retrieved, or evicted in the current framework. As video streams become longer or multi-turn interactions accumulate, historical KV states may still dominate memory consumption and become a bottleneck~\citep{zeng2025streamforest, zhang2024flashvstream}. In future work, we aim to study joint frame selection and KV-cache management toward better controlling memory growth while preserving query-relevant visual evidence.

\section{Conclusion}
This paper presented CoFiE, a coarse-to-fine evidence selection framework for efficient streaming video understanding. CoFiE first uses query-agnostic visual novelty to keep a high-recall set of candidate frames before visual encoding, then applies query-specific refinement during multimodal prefill to select evidence frames for generation. This design reduces redundant visual computation early while preserving the question-relevant evidence needed for accurate responses. Experiments on streaming and long-video benchmarks show that CoFiE achieves state-of-the-art accuracy--efficiency trade-offs, indicating that frame-level evidence selection is an effective path toward low-latency streaming VLLMs.

\section*{Acknowledgments}
This work was supported by the National Natural Science Foundation of China (grant No. 62350710797). Additionally, the authors gratefully acknowledge the support of the NSFC Excellent Young Scientists Fund Program (Overseas).


\bibliography{anthology}

@INPROCEEDINGS{histogram2013,
  author={Janwe, Nitin J. and Bhoyar, Kishor K.},
  booktitle={2013 IEEE Second International Conference on Image Information Processing (ICIIP-2013)}, 
  title={Video shot boundary detection based on JND color histogram}, 
  year={2013},
  volume={},
  number={},
  pages={476-480},
  doi={10.1109/ICIIP.2013.6707637}}

@misc{google2024gemini15,
  title = {Gemini 1.5: Unlocking Multimodal Understanding across Millions of Tokens of Context},
  author = {{Gemini Team, Google}},
  year = {2024},
  publisher = {arXiv},
  doi = {10.48550/ARXIV.2403.05530},
  url = {https://arxiv.org/abs/2403.05530},
}

@misc{openai2024gpt4o,
  title = {{GPT-4o} System Card},
  author = {{OpenAI}},
  year = {2024},
  publisher = {arXiv},
  doi = {10.48550/ARXIV.2410.21276},
  url = {https://arxiv.org/abs/2410.21276},
}

@misc{anthropic2024claude35sonnet,
  title = {{Claude 3.5 Sonnet} Model Card Addendum},
  author = {{Anthropic}},
  year = {2024},
  publisher = {Anthropic},
  url = {https://www-cdn.anthropic.com/fed9cc193a14b84131812372d8d5857f8f304c52/Model_Card_Claude_3_Addendum.pdf},
}

@misc{yao2024minicpmv,
  title = {{MiniCPM-V}: A {GPT-4V} Level {MLLM} on Your Phone},
  author = {Yao, Yuan and Yu, Tianyu and Zhang, Ao and Wang, Chongyi and Cui, Junbo and Zhu, Hongji and Cai, Tianchi and Li, Haoyu and Zhao, Weilin and He, Zhihui and Chen, Qianyu and Zhou, Huarong and Zou, Zhensheng and Zhang, Haoye and Hu, Shengding and Zheng, Zhi and Zhou, Jie and Cai, Jie and Han, Xu and Zeng, Guoyang and Li, Dahai and Liu, Zhiyuan and Sun, Maosong},
  year = {2024},
  publisher = {arXiv},
  doi = {10.48550/ARXIV.2408.01800},
  url = {https://arxiv.org/abs/2408.01800},
}

@misc{opengvlab2024internvl2,
  title = {{InternVL2}: Better than the Best---Expanding Performance Boundaries of Open-Source Multimodal Models with the Progressive Scaling Strategy},
  author = {{OpenGVLab Team}},
  year = {2024},
  howpublished = {\url{https://internvl.github.io/blog/2024-07-02-InternVL-2.0/}},
  publisher = {OpenGVLab}
}

@inproceedings{lin2024vila,
  title = {{VILA}: On Pre-training for Visual Language Models},
  author = {Lin, Ji and Yin, Hongxu and Ping, Wei and Lu, Yao and Molchanov, Pavlo and Tao, Andrew and Mao, Huizi and Kautz, Jan and Shoeybi, Mohammad and Han, Song},
  booktitle = {Proceedings of the IEEE/CVF Conference on Computer Vision and Pattern Recognition (CVPR)},
  month = jun,
  year = {2024},
}

@misc{cheng2024videollama2,
  title = {{VideoLLaMA} 2: Advancing Spatial-Temporal Modeling and Audio Understanding in Video-LLMs},
  author = {Cheng, Zesen and Leng, Sicong and Zhang, Hang and Xin, Yifei and Li, Xin and Chen, Guanzheng and Zhu, Yongxin and Zhang, Wenqi and Luo, Ziyang and Zhao, Deli and Bing, Lidong},
  year = {2024},
  publisher = {arXiv},
  doi = {10.48550/ARXIV.2406.07476},
  url = {https://arxiv.org/abs/2406.07476},
}

@misc{bai2025qwen25vl,
  title = {Qwen2.5-VL Technical Report},
  author = {Bai, Shuai and Chen, Keqin and Liu, Xuejing and Wang, Jialin and Ge, Wenbin and Song, Sibo and Dang, Kai and Wang, Peng and Wang, Shijie and Tang, Jun and Zhong, Humen and Zhu, Yuanzhi and Yang, Mingkun and Li, Zhaohai and Wan, Jianqiang and Wang, Pengfei and Ding, Wei and Fu, Zheren and Xu, Yiheng and Ye, Jiabo and Zhang, Xi and Xie, Tianbao and Cheng, Zesen and Zhang, Hang and Yang, Zhibo and Xu, Haiyang and Lin, Junyang},
  year = {2025},
  publisher = {arXiv},
  doi = {10.48550/ARXIV.2502.13923},
  url = {https://arxiv.org/abs/2502.13923},
}

@misc{bai2025qwen3vl,
  title = {Qwen3-VL Technical Report},
  author = {Bai, Shuai and Cai, Yuxuan and Chen, Ruizhe and Chen, Keqin and Chen, Xionghui and Cheng, Zesen and Deng, Lianghao and Ding, Wei and Gao, Chang and Ge, Chunjiang and Ge, Wenbin and Guo, Zhifang and Huang, Qidong and Huang, Jie and Huang, Fei and Hui, Binyuan and Jiang, Shutong and Li, Zhaohai and Li, Mingsheng and Li, Mei and Li, Kaixin and Lin, Zicheng and Lin, Junyang and Liu, Xuejing and Liu, Jiawei and Liu, Chenglong and Liu, Yang and Liu, Dayiheng and Liu, Shixuan and Lu, Dunjie and Luo, Ruilin and Lv, Chenxu and Men, Rui and Meng, Lingchen and Ren, Xuancheng and Ren, Xingzhang and Song, Sibo and Sun, Yuchong and Tang, Jun and Tu, Jianhong and Wan, Jianqiang and Wang, Peng and Wang, Pengfei and Wang, Qiuyue and Wang, Yuxuan and Xie, Tianbao and Xu, Yiheng and Xu, Haiyang and Xu, Jin and Yang, Zhibo and Yang, Mingkun and Yang, Jianxin and Yang, An and Yu, Bowen and Zhang, Fei and Zhang, Hang and Zhang, Xi and Zheng, Bo and Zhong, Humen and Zhou, Jingren and Zhou, Fan and Zhou, Jing and Zhu, Yuanzhi and Zhu, Ke},
  year = {2025},
  publisher = {arXiv},
  doi = {10.48550/ARXIV.2511.21631},
  url = {https://arxiv.org/abs/2511.21631},
}

@inproceedings{
dao2024flashattention,
title={FlashAttention-2: Faster Attention with Better Parallelism and Work Partitioning},
author={Tri Dao},
booktitle={The Twelfth International Conference on Learning Representations},
year={2024},
url={https://openreview.net/forum?id=mZn2Xyh9Ec}
}

@inproceedings{lin2024streamingbench,
  title = {StreamingBench: Assessing the Gap for MLLMs to Achieve Streaming Video Understanding},
  author = {Lin, Junming and Fang, Zheng and Chen, Chi and Cheng, Haoxuan and Wan, Zihao and Luo, Fuwen and Wang, Ziyue and Li, Peng and Liu, Yang and Sun, Maosong},
  booktitle = {Proceedings of the IEEE International Conference on Acoustics, Speech and Signal Processing (ICASSP)},
  pages = {12147--12151},
  year = {2026},
}

@inproceedings{li2025ovobench,
  title = {OVO-Bench: How Far is Your Video-LLMs from Real-World Online Video Understanding?},
  author = {Niu, Junbo and Li, Yifei and Miao, Ziyang and Ge, Chunjiang and Zhou, Yuanhang and He, Qihao and Dong, Xiaoyi and Duan, Haodong and Ding, Shuangrui and Qian, Rui and Zhang, Pan and Zang, Yuhang and Cao, Yuhang and He, Conghui and Wang, Jiaqi},
  booktitle = {Proceedings of the IEEE/CVF Conference on Computer Vision and Pattern Recognition (CVPR)},
  pages = {18902--18913},
  month = jun,
  year = {2025},
}

@inproceedings{zhou2025mlvu,
  title = {{MLVU}: Benchmarking Multi-task Long Video Understanding},
  author = {Zhou, Junjie and Shu, Yan and Zhao, Bo and Wu, Boya and Liang, Zhengyang and Xiao, Shitao and Qin, Minghao and Yang, Xi and Xiong, Yongping and Zhang, Bo and Huang, Tiejun and Liu, Zheng},
  booktitle = {Proceedings of the IEEE/CVF Conference on Computer Vision and Pattern Recognition (CVPR)},
  pages = {13691--13701},
  month = jun,
  year = {2025},
}

@inproceedings{wu2024longvideobench,
  title = {{LongVideoBench}: A Benchmark for Long-context Interleaved Video-Language Understanding},
  author = {Wu, Haoning and Li, Dongxu and Chen, Bei and Li, Junnan},
  booktitle = {Advances in Neural Information Processing Systems},
  volume = {37},
  year = {2024},
}

@inproceedings{li2024mvbench,
  title = {{MVBench}: A Comprehensive Multi-modal Video Understanding Benchmark},
  author = {Li, Kunchang and Wang, Yali and He, Yinan and Li, Yizhuo and Wang, Yi and Liu, Yi and Wang, Zun and Xu, Jilan and Chen, Guo and Luo, Ping and Wang, Limin and Qiao, Yu},
  booktitle = {Proceedings of the IEEE/CVF Conference on Computer Vision and Pattern Recognition (CVPR)},
  pages = {22195--22206},
  month = jun,
  year = {2024},
}

@inproceedings{fu2025videomme,
  title = {Video-MME: The First-Ever Comprehensive Evaluation Benchmark of Multi-modal LLMs in Video Analysis},
  author = {Fu, Chaoyou and Dai, Yuhan and Luo, Yongdong and Li, Lei and Ren, Shuhuai and Zhang, Renrui and Wang, Zihan and Zhou, Chenyu and Shen, Yunhang and Zhang, Mengdan and Chen, Peixian and Li, Yanwei and Lin, Shaohui and Zhao, Sirui and Li, Ke and Xu, Tong and Zheng, Xiawu and Chen, Enhong and Shan, Caifeng and He, Ran and Sun, Xing},
  booktitle = {Proceedings of the IEEE/CVF Conference on Computer Vision and Pattern Recognition (CVPR)},
  pages = {24108--24118},
  month = jun,
  year = {2025},
}

@inproceedings{chen2024videollmonline,
  title = {VideoLLM-online: Online Video Large Language Model for Streaming Video},
  author = {Chen, Joya and Lv, Zhaoyang and Wu, Shiwei and Lin, Kevin Qinghong and Song, Chenan and Gao, Difei and Liu, Jia-Wei and Gao, Ziteng and Mao, Dongxing and Shou, Mike Zheng},
  booktitle = {Proceedings of the IEEE/CVF Conference on Computer Vision and Pattern Recognition (CVPR)},
  pages = {18407--18418},
  month = jun,
  year = {2024},
}

@inproceedings{zhang2024flashvstream,
  title = {Flash-VStream: Efficient Real-Time Understanding for Long Video Streams},
  author = {Zhang, Haoji and Wang, Yiqin and Tang, Yansong and Liu, Yong and Feng, Jiashi and Jin, Xiaojie},
  booktitle = {Proceedings of the IEEE/CVF International Conference on Computer Vision (ICCV)},
  pages = {21059--21069},
  month = oct,
  year = {2025},
}

@inproceedings{qian2025dispider,
  title = {Dispider: Enabling Video LLMs with Active Real-Time Interaction via Disentangled Perception, Decision, and Reaction},
  author = {Qian, Rui and Ding, Shuangrui and Dong, Xiaoyi and Zhang, Pan and Zang, Yuhang and Cao, Yuhang and Lin, Dahua and Wang, Jiaqi},
  booktitle = {Proceedings of the IEEE/CVF Conference on Computer Vision and Pattern Recognition (CVPR)},
  pages = {24045--24055},
  month = jun,
  year = {2025},
}

@inproceedings{qian2024videostreaming,
  title = {Streaming Long Video Understanding with Large Language Models},
  author = {Qian, Rui and Dong, Xiaoyi and Zhang, Pan and Zang, Yuhang and Ding, Shuangrui and Lin, Dahua and Wang, Jiaqi},
  booktitle = {Advances in Neural Information Processing Systems},
  volume = {37},
  pages = {119336--119360},
  year = {2024},
}

@inproceedings{yao2025timechatonline,
  title={Timechat-online: 80\% visual tokens are naturally redundant in streaming videos},
  author={Yao, Linli and Li, Yicheng and Wei, Yuancheng and Li, Lei and Ren, Shuhuai and Liu, Yuanxin and Ouyang, Kun and Wang, Lean and Li, Shicheng and Li, Sida and Kong, Lingpeng and Liu, Qi and Zhang, Yuanxing and Sun, Xu},
  booktitle={Proceedings of the 33rd ACM International Conference on Multimedia},
  pages={10807--10816},
  year={2025}
}

@inproceedings{fu2025vispeak,
  title = {ViSpeak: Visual Instruction Feedback in Streaming Videos},
  author = {Fu, Shenghao and Yang, Qize and Li, Yuan-Ming and Peng, Yi-Xing and Lin, Kun-Yu and Wei, Xihan and Hu, Jian-Fang and Xie, Xiaohua and Zheng, Wei-Shi},
  booktitle = {Proceedings of the IEEE/CVF International Conference on Computer Vision (ICCV)},
  pages = {21778--21788},
  month = oct,
  year = {2025},
}

@inproceedings{zeng2025streamforest,
  title = {StreamForest: Efficient Online Video Understanding with Persistent Event Memory},
  author = {Zeng, Xiangyu and Qiu, Kefan and Zhang, Qingyu and Li, Xinhao and Wang, Jing and Li, Jiaxin and Yan, Ziang and Tian, Kun and Tian, Meng and Zhao, Xinhai and Wang, Yi and Wang, Limin},
  booktitle = {Advances in Neural Information Processing Systems},
  volume = {38},
  year = {2025},
}

@inproceedings{zhang2025lmmeval,
    title = "{LMM}s-Eval: Reality Check on the Evaluation of Large Multimodal Models",
    author = "Zhang, Kaichen  and
      Li, Bo  and
      Zhang, Peiyuan  and
      Pu, Fanyi  and
      Cahyono, Joshua Adrian  and
      Hu, Kairui  and
      Liu, Shuai  and
      Zhang, Yuanhan  and
      Yang, Jingkang  and
      Li, Chunyuan  and
      Liu, Ziwei",
    editor = "Chiruzzo, Luis  and
      Ritter, Alan  and
      Wang, Lu",
    booktitle = "Findings of the Association for Computational Linguistics: NAACL 2025",
    month = apr,
    year = "2025",
    address = "Albuquerque, New Mexico",
    publisher = "Association for Computational Linguistics",
    url = "https://aclanthology.org/2025.findings-naacl.51/",
    doi = "10.18653/v1/2025.findings-naacl.51",
    pages = "881--916",
    ISBN = "979-8-89176-195-7",
}

@inproceedings{chen2024fastv,
  title={An Image is Worth 1/2 Tokens After Layer 2: Plug-and-Play Inference Acceleration for Large Vision-Language Models},
  author={Chen, Liang and Zhao, Haozhe and Liu, Tianyu and Bai, Shuai and Lin, Junyang and Zhou, Chang and Chang, Baobao},
  booktitle={European Conference on Computer Vision},
  pages={19--35},
  year={2024}
}

@inproceedings{huang2024prunevid,
  title={Prunevid: Visual token pruning for efficient video large language models},
  author={Huang, Xiaohu and Zhou, Hao and Han, Kai},
  booktitle={Findings of the Association for Computational Linguistics: ACL 2025},
  pages={19959--19973},
  year={2025}
}

@misc{zeng2025glimpseprune,
  title = {A Glimpse to Compress: Dynamic Visual Token Pruning for Large Vision-Language Models},
  author = {Zeng, Quan-Sheng and Li, Yunheng and Wang, Qilong and Jiang, Peng-Tao and Wu, Zuxuan and Cheng, Ming-Ming and Hou, Qibin},
  year = {2025},
  publisher = {arXiv},
  doi = {10.48550/ARXIV.2508.01548},
  url = {https://arxiv.org/abs/2508.01548},
}

@inproceedings{chen2023internvl,
  title={Internvl: Scaling up vision foundation models and aligning for generic visual-linguistic tasks},
  author={Chen, Zhe and Wu, Jiannan and Wang, Wenhai and Su, Weijie and Chen, Guo and Xing, Sen and Zhong, Muyan and Zhang, Qinglong and Zhu, Xizhou and Lu, Lewei and Li, Bin and Luo, Ping and Lu, Tong and Qiao, Yu  and Dai, Jifeng},
  booktitle={Proceedings of the IEEE/CVF conference on computer vision and pattern recognition},
  pages={24185--24198},
  year={2024}
}

@article{li2024llava,
  title={Llava-onevision: Easy visual task transfer},
  author={Li, Bo and Zhang, Yuanhan and Guo, Dong and Zhang, Renrui  and Li, Feng and Zhang , Hao and Zhang, Kaichen and Zhang, Peiyuan  and Li, Yanwei and Liu, Ziwei and Li, Chunyuan},
  journal={arXiv preprint arXiv:2408.03326},
  year={2024}
}

@article{wang2024qwen2vl,
  title={Qwen2-vl: Enhancing vision-language model's perception of the world at any resolution},
  author={Wang, Peng  and Bai, Shuai and Tan, Sinan and Wang, Shijie and Fan, Zhihao and Bai, Jinze and Chen, Keqin and Liu, Xuejing and Wang, Jialin and Ge, Wenbin and Fan, Yang and Dang, Kai and Du, Mengfei and Ren, Xuancheng and Men, Rui and Liu, Dayiheng and Zhou, Chang  and Zhou, Jingren and Lin, Junyang},
  journal={arXiv preprint arXiv:2409.12191},
  year={2024}
}

@inproceedings{lipton1998moving,
  title     = {Moving Target Classification and Tracking from Real-Time Video},
  author    = {Lipton, Alan J. and Fujiyoshi, Hironobu and Patil, Raju S.},
  booktitle = {Proceedings of the 1998 IEEE Workshop on Applications of Computer Vision},
  pages     = {8--14},
  year      = {1998},
  publisher = {IEEE},
  doi       = {10.1109/ACV.1998.732851}
}

@mastersthesis{zauner2010implementation,
  title  = {Implementation and Benchmarking of Perceptual Image Hash Functions},
  author = {Zauner, Christoph},
  school = {Upper Austria University of Applied Sciences},
  year   = {2010}
}

@article{wang2004image,
  title   = {Image Quality Assessment: From Error Visibility to Structural Similarity},
  author  = {Wang, Zhou and Bovik, Alan C. and Sheikh, Hamid R. and Simoncelli, Eero P.},
  journal = {IEEE Transactions on Image Processing},
  volume  = {13},
  number  = {4},
  pages   = {600--612},
  year    = {2004},
  doi     = {10.1109/TIP.2003.819861}
}

@article{horn1981determining,
  title   = {Determining Optical Flow},
  author  = {Horn, Berthold K. P. and Schunck, Brian G.},
  journal = {Artificial Intelligence},
  volume  = {17},
  number  = {1--3},
  pages   = {185--203},
  year    = {1981},
  doi     = {10.1016/0004-3702(81)90024-2}
}

@article{zhang1993automatic,
  title   = {Automatic Partitioning of Full-Motion Video},
  author  = {Zhang, HongJiang and Kankanhalli, Atreyi and Smoliar, Stephen W.},
  journal = {Multimedia Systems},
  volume  = {1},
  number  = {1},
  pages   = {10--28},
  year    = {1993},
  doi     = {10.1007/BF01210504}
}

@article{shen2024longvu,
  author ={Shen, Xiaoqian and Xiong, Yunyang and Zhao, Changsheng and Wu, Lemeng and Chen, Jun and Zhu, Chenchen and Liu, Zechun and Xiao, Fanyi and Varadarajan, Balakrishnan and Bordes, Florian and Liu, Zhuang and Xu, Hu and J. Kim, Hyunwoo and Soran, Bilge and Krishnamoorthi, Raghuraman and Elhoseiny, Mohamed and Chandra, Vikas},
  title = {LongVU: Spatiotemporal Adaptive Compression for Long Video-Language Understanding},
  journal = {arXiv preprint arXiv:2410.17434},
  year = {2024},
}

\newpage
\appendix
\section{Implementation Details}
\label{sec:appendix-impl}

We implement \method{} on Qwen3-VL-8B-Instruct. The Vision Transformer (ViT) encoder remains frozen during training, while the multimodal projector and large language model (LLM) are fine-tuned to help the base model leverage sparse visual evidence. Fine-tuning and evaluation use NVIDIA Pro6000-96GB GPUs with FlashAttention-2~\citep{dao2024flashattention} and the LMMs-Eval framework~\citep{zhang2025lmmeval}. The NGFF module is integrated into the video processor before visual feature encoding, quantifying frame novelty with 32-bin grayscale histograms and removing temporally redundant frames before the computationally intensive vision encoder. The QER module further refines the retained frame candidates during LLM pre-filling by scoring query-frame relevance using the last-layer text-to-vision attention map. The final frame retention ratio is set to 30\%--50\%, according to the target compression budget.

\section{Additional Experimental Details}
\label{sec:appendix-exp}

\subsection{Frame Dropping versus Token Dropping}

\begin{table}[h]
\centering
\scriptsize
\setlength{\tabcolsep}{5.8pt}
\renewcommand{\arraystretch}{1.12}
\resizebox{\columnwidth}{!}{
\begin{tabular}{lccc!{\vrule width 0.3pt}c}
\toprule
\textbf{Method} & \textbf{Drop (\%)} & \textbf{Avg. Infer. time} & \textbf{Speedup} & \textbf{Acc. ($\Delta$Acc.)} \\
\midrule
Qwen3-VL-8B & 0 & 2.68s & 1.00$\times$ & 65.03  \\
\midrule
\rowcolor{black!8}
\multicolumn{5}{c}{\textit{Token-level dropping}} \\
\multirow{2}{*}{TimeChat-Online} & 50 & 2.31s & 1.16$\times$ & 60.48  \\
 & 80 & 2.27s & 1.18$\times$ & 57.12 (-3.36) \\
\midrule
\rowcolor{black!8}
\multicolumn{5}{c}{\textit{Frame-level dropping}} \\
\multirow{2}{*}{\method{}} & 50 & 1.81s & 1.48$\times$ & \textbf{68.72}  \\
 & 80 & \textbf{1.06s} & \textbf{2.54$\times$} & 66.00 (-2.72) \\
\bottomrule
\end{tabular}
}
\caption{Frame-level versus token-level dropping on OvO-Bench Real-Time Visual Perception category. Average inference time is computed over videos of varying lengths. Speedup is measured relative to the unpruned Qwen3-VL-8B baseline.}
\label{tab:frame-vs-token}
\end{table}

Table~\ref{tab:frame-vs-token} compares frame-level and token-level dropping on OvO-Bench under the same Qwen3-VL-8B backbone.
At the 50\% drop ratio, TimeChat-Online reduces inference time from 2.68s to 2.31s, yielding only a 1.16$\times$ speedup while degrading accuracy to 60.48.
In contrast, \method{} reduces inference time to 1.81s and improves accuracy to 68.72, indicating that redundant frames can be removed before visual encoding without discarding task-critical visual evidence.
The gap becomes larger at the 80\% drop ratio.
Token-level dropping achieves only a 1.18$\times$ speedup and retains 87.8\% of the unpruned accuracy, whereas \method{} reaches a 2.54$\times$ speedup while preserving 101.5\% of the no-drop performance.
These results show that frame-level dropping is more effective for end-to-end acceleration because it removes redundant visual computation before the vision encoder, while token-level dropping only shortens the downstream multimodal context after the dominant front-end cost has already been paid.

\subsection{NGFF Hyperparameter Sensitivity}

We evaluate the sensitivity of NGFF to three hyperparameters: the minimum number of retained frames $M_{\min}$, the maximum forced-retention interval $L_{\max}$, and the histogram-difference threshold $\tau_h$. Experiments use Qwen3-VL-8B on a subset of StreamingBench. Table~\ref{tab:ngff-sensitivity} varies one hyperparameter at a time while holding the other two fixed.

\begin{table}[t]
\centering
\scriptsize
\setlength{\tabcolsep}{3.0pt}
\renewcommand{\arraystretch}{1.05}
\resizebox{\columnwidth}{!}{
\begin{tabular}{cccccc}
\toprule
$M_{\min}$ & $L_{\max}$ & $\tau_h$ & Acc. $\uparrow$ & Drop (\%) & Time (s) $\downarrow$ \\
\midrule
2 & 30 & 0.30 & 65.20 & 82.29 & 119.84 \\
4 & 30 & 0.30 & 66.00 & 82.14 & 120.02 \\
8 & 30 & 0.30 & 66.00 & 81.60 & 121.25 \\
\midrule
4 & 15 & 0.30 & 67.60 & 80.44 & 125.58 \\
4 & 30 & 0.30 & 66.00 & 82.14 & 120.02 \\
4 & 45 & 0.30 & 66.80 & 82.55 & 118.66 \\
\midrule
4 & 30 & 0.15 & 68.40 & 61.60 & 209.28 \\
4 & 30 & 0.30 & 66.00 & 82.14 & 120.02 \\
4 & 30 & 0.45 & 70.00 & 89.44 & 91.81 \\
4 & 30 & 0.60 & 68.40 & 92.44 & 80.71 \\
4 & 30 & 0.75 & 68.80 & 93.98 & 75.30 \\
\bottomrule
\end{tabular}
}
\caption{Sensitivity of NGFF hyperparameters on a StreamingBench subset.}
\label{tab:ngff-sensitivity}
\end{table}

As shown in Table~\ref{tab:ngff-sensitivity}, performance varies by less than 1\% across the evaluated $M_{\min}$ and $L_{\max}$ ranges. $M_{\min}$ acts as a lower bound that becomes active only for very short videos with frequent visual changes, while $L_{\max}$ prevents long static intervals from becoming overly sparse. The threshold $\tau_h$ directly controls the number of candidate frames and therefore exposes the retention--latency trade-off. The method is robust to reasonable hyperparameter choices within the tested ranges.

\subsection{Allocation between NGFF and QER}

We further study how a fixed total frame-drop budget should be allocated between the query-agnostic NGFF stage and the query-specific QER stage. Experiments use Qwen3-VL-8B on a subset of Video-MME.

\begin{table}[t]
\centering
\scriptsize
\setlength{\tabcolsep}{2.4pt}
\renewcommand{\arraystretch}{1.05}
\resizebox{\columnwidth}{!}{
\begin{tabular}{ccccc}
\toprule
Total drop & NGFF drop & QER drop & Time $\downarrow$ & Acc. $\uparrow$ \\
\midrule
0\% & 0\% & 0\% & 15m35s & 71.1 \\
80\% & 0\% & 80\% & 13m32s & 71.9 \\
80\% & 20\% & 75\% & 9m51s & 72.6 \\
80\% & 35\% & 70\% & 7m35s & 68.9 \\
80\% & 50\% & 60\% & 5m37s & 70.4 \\
80\% & 70\% & 33.4\% & 3m06s & 65.2 \\
80\% & 80\% & 0\% & 2m06s & 59.3 \\
\bottomrule
\end{tabular}
}
\caption{Allocation of an approximately fixed total drop ratio between NGFF and QER on a Video-MME subset.}
\label{tab:drop-allocation}
\end{table}

As shown in Table~\ref{tab:drop-allocation}, the allocations form different accuracy--efficiency operating points. Assigning more of the budget to NGFF reduces visual encoding cost and improves latency, whereas retaining more candidates for QER better preserves question-relevant evidence. The 50\% NGFF / 60\% QER split offers a favorable efficiency--accuracy balance, while the 20\% NGFF / 75\% QER split favors accuracy. These results indicate that the two stages provide complementary rather than interchangeable capabilities.

\subsection{Comparison with Model-Based Filtering}

To compare NGFF with a model-based filtering criterion, we use CLIP ViT-B/32 to extract frame features and retain frames with lower adjacent-frame similarity. We evaluate this variant with Qwen3-VL-8B on a subset of OvO-Bench.

\begin{table}[t]
\centering
\scriptsize
\setlength{\tabcolsep}{2.0pt}
\renewcommand{\arraystretch}{1.05}
\resizebox{\columnwidth}{!}{
\begin{tabular}{lcccc}
\toprule
NGFF signal & Total drop & NGFF drop & QER drop & Acc. / Time \\
\midrule
Baseline & 0\% & 0\% & 0\% & 64.51 / 275.02s \\
Histogram difference & 20\% & 20\% & 0\% & \textbf{66.97} / \textbf{194.60s} \\
CLIP similarity & 20\% & 20\% & 0\% & 65.78 / 275.51s \\
Histogram difference & 60\% & 20\% & 50\% & \textbf{66.97} / \textbf{152.24s} \\
CLIP similarity & 60\% & 20\% & 50\% & 65.78 / 212.98s \\
\bottomrule
\end{tabular}
}
\caption{Histogram-difference and CLIP-based filtering on an OvO-Bench subset.}
\label{tab:clip-filtering}
\end{table}

As shown in Table~\ref{tab:clip-filtering}, histogram-difference filtering achieves higher accuracy and lower latency than CLIP-based filtering. Although CLIP provides semantic visual features, the additional feature-extraction cost largely negates the latency reduction from filtering. The histogram signal thus achieves a better end-to-end accuracy--latency trade-off at the pre-encoder stage.

\subsection{Attention-Layer Selection for QER}

We evaluate the attention layer used by QER at a fixed 60\% frame drop ratio with Qwen3-VL-8B on a subset of StreamingBench.

\begin{table}[t]
\centering
\scriptsize
\setlength{\tabcolsep}{7pt}
\renewcommand{\arraystretch}{1.05}
\begin{tabular}{cc}
\toprule
Layer & Acc. $\uparrow$ \\
\midrule
0 & 58.80 \\
4 & 59.60 \\
8 & 60.80 \\
12 & 62.00 \\
18 & 67.20 \\
24 & \textbf{68.00} \\
30 & 66.80 \\
35 & 67.20 \\
\bottomrule
\end{tabular}
\caption{Layer-wise ablation for QER at a 60\% drop ratio on a StreamingBench subset.}
\label{tab:qer-layer}
\end{table}

As shown in Table~\ref{tab:qer-layer}, middle-to-late layers consistently outperform early layers, and Layer 24 yields the highest accuracy in this subset. We adopt the final layer in the main method because its attention signal is directly associated with output prediction and because it generalizes across architectures without requiring layer-index tuning. The consistently strong results from Layers 18--35 indicate that QER is robust to the layer choice within this range.

\subsection{Preservation of Answer-Relevant Evidence}

NGFF compares each frame with both the preceding frame and the most recently retained frame, always retains the first and last frames, limits the maximum retention interval, and adds uniformly sampled frames when fewer than $M_{\min}$ candidates remain. To directly evaluate whether these safeguards preserve answer-relevant content, we use query-conditioned attention obtained by applying QER to the full, unfiltered video. The question and its four answer options are used as textual queries. We define \textit{Attention Preservation Ratio} as the fraction of full-video answer-relevant attention mass assigned to the frames retained by NGFF, and \textit{Top-Evidence Recall@20\%} as the fraction of the top 20\% most answer-relevant frames retained after coarse filtering. Experiments use Qwen3-VL-8B on a subset of StreamingBench.

\begin{table}[t]
\centering
\scriptsize
\setlength{\tabcolsep}{2.5pt}
\renewcommand{\arraystretch}{1.05}
\resizebox{\columnwidth}{!}{
\begin{tabular}{lccc}
\toprule
Method & Drop & Attention preservation $\uparrow$ & Recall@20\% $\uparrow$ \\
\midrule
Uniform sampling & 20\% & 62.74 & 63.24 \\
Random sampling & 20\% & 65.18 & 65.44 \\
\textbf{NGFF} & 20\% & \textbf{69.69} & \textbf{69.58} \\
\midrule
Uniform sampling & 50\% & 5.28 & 4.80 \\
Random sampling & 50\% & 28.45 & 27.83 \\
\textbf{NGFF} & 50\% & \textbf{36.83} & \textbf{39.55} \\
\bottomrule
\end{tabular}
}
\caption{Preservation of answer-relevant evidence by coarse filtering on a StreamingBench subset.}
\label{tab:evidence-preservation}
\end{table}

As shown in Table~\ref{tab:evidence-preservation}, at both drop ratios, NGFF retains more answer-relevant attention and more top-evidence frames than uniform or random sampling. The gap widens under the more aggressive 50\% drop setting, which is consistent with NGFF's role as a coarse candidate selector preceding query-specific refinement.

\subsection{Generalization across VLLM Backbones}

We additionally implement CoFiE on InternVL3.5-8B, a VLLM from a different model family, and evaluate it on the same subset of StreamingBench used for this cross-backbone comparison.

\begin{table}[t]
\centering
\scriptsize
\setlength{\tabcolsep}{2.0pt}
\renewcommand{\arraystretch}{1.05}
\resizebox{\columnwidth}{!}{
\begin{tabular}{lcccc}
\toprule
Backbone & Total drop & NGFF drop & QER drop & Acc. / Time \\
\midrule
Qwen3-VL-8B & 0\% & 0\% & 0\% & 66.80 / 514.68s \\
Qwen3-VL-8B & 50\% & 20\% & 37.5\% & 67.60 / 312.19s \\
InternVL3.5-8B & 0\% & 0\% & 0\% & 64.00 / 441.50s \\
InternVL3.5-8B & 50\% & 20\% & 37.5\% & 66.40 / 383.28s \\
\bottomrule
\end{tabular}
}
\caption{Cross-backbone evaluation on a StreamingBench subset.}
\label{tab:cross-backbone}
\end{table}

As shown in Table~\ref{tab:cross-backbone}, at a 50\% total drop ratio, CoFiE improves Qwen3-VL-8B accuracy by 0.80 percentage points while providing a 1.65$\times$ speedup. On InternVL3.5-8B, it improves accuracy by 2.40 percentage points while providing a 1.15$\times$ speedup. These results show that the coarse-to-fine selection strategy transfers across the two VLLM families; the varying speedups arise from differences in how each backbone distributes latency across visual encoding, prefill, and decoding.

\end{document}